\documentclass[10pt,twocolumn,letterpaper]{article}

\usepackage{cvpr}      

\usepackage{bm}
\definecolor{cvprblue}{rgb}{0.21,0.49,0.74}
\definecolor{tea}{RGB}{238, 238, 238} 
\definecolor{Gray}{gray}{0.9}
\usepackage[pagebackref,breaklinks,colorlinks,citecolor=cvprblue]{hyperref}
\usepackage{color, colortbl}
\usepackage{multirow}
\usepackage{booktabs} 
\usepackage[T1]{fontenc}
\usepackage{hhline}
\usepackage{bbding}
\usepackage[ruled,linesnumbered]{algorithm2e}
\usepackage{caption, subcaption}
 
\def\paperID{****} 
\def\confName{CVPR}
\def\confYear{2026}

\title{GMT: Effective Global Framework for Multi-Camera Multi-Target Tracking }

\author{Yihao Zhen$^{1,3}$, Mingyue Xu$^{1,3}$, Qiang Wang$^{2}$, Baojie Fan$^{4}$, Jiahua Dong$^{5}$, Tinghui Zhao$^{1,3}$, Huijie Fan$^{1,3}$\thanks{Corresponding author}\\
$^{1}$State Key Laboratory of Robotics and Intelligent Systems,\\
Shenyang Institute of Automation, Chinese Academy of Sciences\\
$^{2}$Key Laboratory of Manufacturing Industrial Integrated Automation, Shenyang University\\
$^{3}$University of Chinese Academy of Sciences  $^{4}$Nanjing University of Posts and Telecommunications\\
$^{5}$Mohamed bin Zayed University of Artificial Intelligence\\
{\tt\small {\{fanhuijie,qiaoyu,zhenyihao,zhaotinghui,wangqiang\}@sia.cn, jobfbj@gmail.com}}
}

\begin{document}
\maketitle
\begin{abstract}
Multi-Camera Multi-Target (MCMT) tracking aims to locate and associate the same targets across multiple camera views. Existing methods typically adopt a two-stage framework, involving single-camera tracking followed by inter-camera tracking. However, in this paradigm, multi-view information is used only to recover missed matches in the first stage, providing a limited contribution to overall tracking. To address this issue, we propose GMT, a global MCMT tracking framework that jointly exploits intra-view and inter-view cues for tracking. Specifically, instead of assigning trajectories independently for each view, we integrate the same historical targets across different views as global trajectories, thereby reformulating the two-stage tracking as a unified global-level trajectory-target association process. We introduce a Cross-View Feature Consistency Enhancement (CFCE) module to align visual and spatial features across views, providing a consistent feature space for global trajectory modeling. With these aligned features, the Global Trajectory Association (GTA) module associates new detections with existing global trajectories, enabling direct use of multi-view information. Compared to the two-stage framework, GMT achieves significant improvements on existing datasets, with gains of up to 21.3 percent in CVMA and 17.2 percent in CVIDF1. Furthermore, we introduce VisionTrack, a high-quality, large-scale MCMT dataset providing significantly greater diversity than existing datasets. Our code and dataset will be released.
\end{abstract}
    
\section{Introduction}
\label{sec:intro}

\begin{figure}  
\centering
\includegraphics[width=1\linewidth]{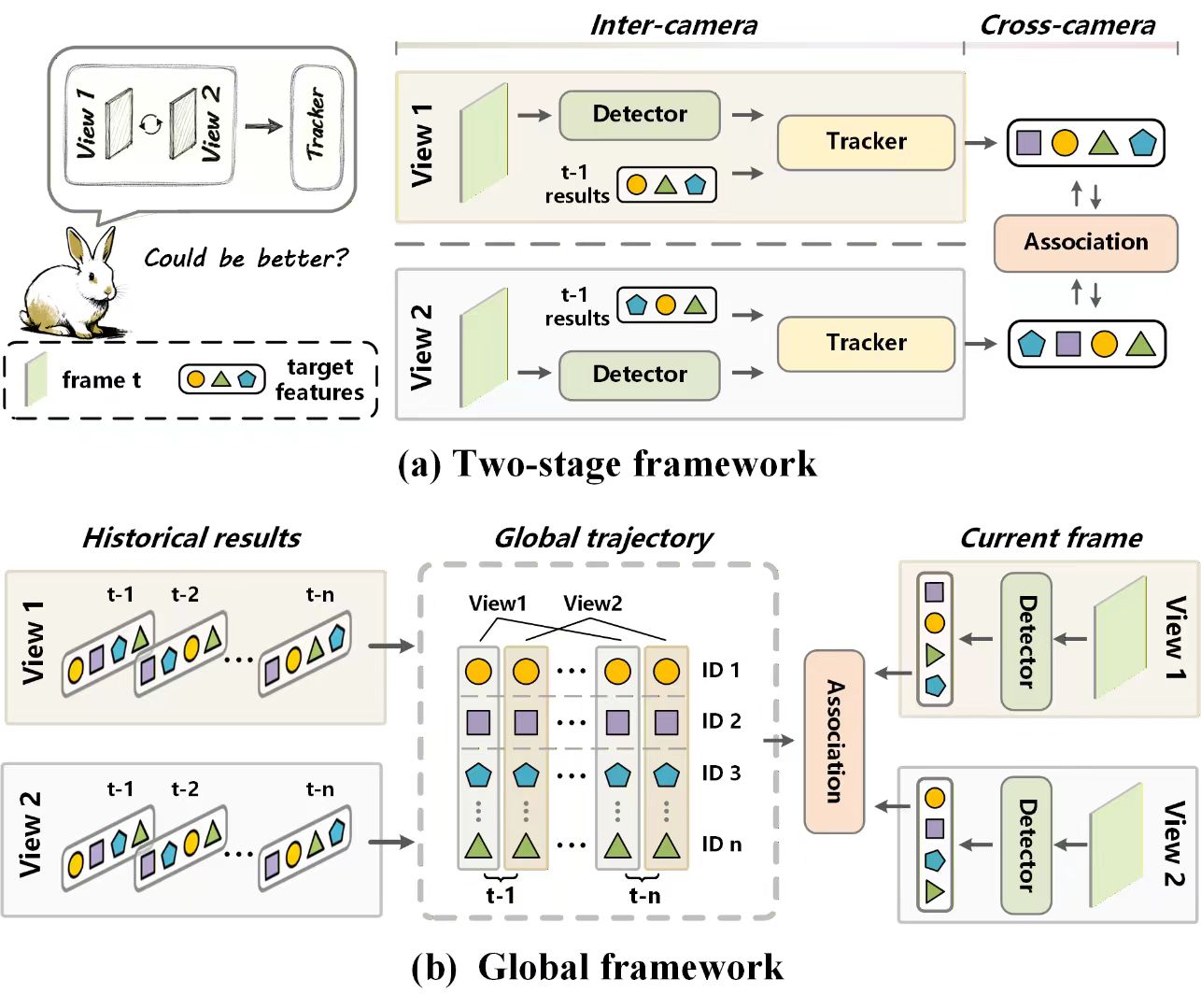}

\caption{\textbf{Two-stage vs Global framework}. Compared with (a) the two-stage paradigm, (b) the proposed global framework directly exploits information from all views for tracking by encoding all historical targets into global trajectories.}

\label{fig:data_vi}
\end{figure}

Multi-Camera Multi-Target (MCMT) \cite{li2019state,li2021visio,Ristani2018FeaturesFM} tracking aims to locate and associate the same targets across multiple overlapping camera views. It has important implications for a number of areas, such as urban reconnaissance \cite{li2020infrared}, crowd behavior analysis \cite{ma2019bayesian}, and traffic scene understanding \cite{shim2021multi}. 
Compared to single-view tracking, an important advantage of MCMT tracking is that incorporating multiple views provides richer visual information for each target. This increased information diversity enables more robust and discriminative target modeling, especially when the target is occluded or undergoes significant appearance changes \cite{OCMC,CVMHT}. 

However, most existing MCMT methods fail to fully exploit this advantage. 
As shown in \cref{fig:data_vi}(a), recent advanced MCMT methods \cite{chavdarova2018wildtrack,he2020multi,nguyen2022lmgp,cheng2023rest} typically adopt the two-stage tracking framework, which employs off-the-shelf multi-object tracking methods to perform Single-Camera Tracking \textbf{(SCT)} independently within each view, followed by Inter-Camera Tracking \textbf{(ICT)} methods to associate the tracking results across views. 
In this paradigm, the SCT stage primarily determines the final tracking results but relies solely on intra-view information. The role of multiple views is limited to correcting targets that were missed in SCT through ICT, contributing only marginally to the overall tracking process. 
Moreover, treating the cross-view association as an independent stage increases the risk of errors when viewpoints vary widely or camera numbers increase.

To address these issues, we propose a novel \textbf{G}lobal MC\textbf{MT} framework, termed \textbf{GMT}. As shown in \cref{fig:data_vi}(b), instead of assigning trajectories independently for each view, GMT directly encodes the same historical targets across different views as global trajectories, thereby converting the MCMT tracking task into a global-level trajectory-target matching task. Compared with the two-stage framework, GMT stands out with two advantages: \textbf{1)} GMT can directly exploit inter-view cues for tracking, as the global trajectories encompass not only temporal context across frames but also diverse visual representations from all views. \textbf{2)} The unnecessary cross-view matching is avoided by directly determining which global trajectory the new targets belong to.

However, encoding features from multiple views into unified trajectories faces a significant challenge: In cross-view tasks, different camera views can be regarded as different domains \cite{Zhang_2022_CVPR,Gait,shiben2}, features captured across different cameras often exhibit discrepancies. To mitigate the negative impact of cross-view inconsistencies on global trajectory modeling, we introduce a \textbf{C}ross-view \textbf{F}eature \textbf{C}onsistency \textbf{E}nhancement \textbf{(CFCE)} module that simultaneously suppresses intra-trajectory discrepancies and enhances inter-trajectory discriminability.
The CFCE module consists of Visual Feature Consistency Enhancement (VFCE) module and Relative Position Consistency Enhancement (RPCE)  module. The VFCE module projects the visual features of targets to a unified trajectory-centric space through metric learning, thereby enforcing appearance coherence along each trajectory. Inspired by the projective consistency across different views, The RPCE module further encodes cross-target relative positional relationships, offering a geometric to visual features.
 The enhanced features are fed into the \textbf{G}lobal \textbf{T}rajectory \textbf{A}ssociation (GTA) module. Within GTA, global trajectories from historical frames are further encoded to enrich the temporal contextual information of trajectory features, while the features of new targets interact with the global trajectories to learn discriminative information across different trajectories. Finally, the GTA module computes the similarity matrix between historical trajectories and new targets and performs trajectory-target matching using the Hungarian algorithm.

Moreover, we construct a novel high-quality, large-scale MCMT tracking dataset, named \textbf{VisionTrack}. Existing datasets are generally confined to a limited number of scenarios, which results in a lack of diversity and relatively small sizes. To address these limitations, VisionTrack encompasses data captured by two moving UAVs across various scenarios, different weather conditions, and at distinct times of the day. It significantly surpasses existing datasets in terms of both scale and diversity.

Our main contributions are summarized as follows:

\begin{itemize}
    \item We propose a novel global MCMT framework, \textbf{GMT}, which could more effectively leverage the advantage of multiple views by encoding the same targets from different views into global trajectories and performing global-level trajectory-target  association. 

    \item We propose a \textbf{CFCE} strategy to align the features of targets within each trajectory, along with a \textbf{GTA} module that enables the trajectory encoding and trajectory-target interaction.
    \item We construct a novel high-quality, large-scale MCMT dataset \textbf{VisionTrack}, which significantly surpasses existing datasets in terms of scale and diversity.
\end{itemize}

\begin{figure*}  
\centering
\includegraphics[width=1\linewidth]{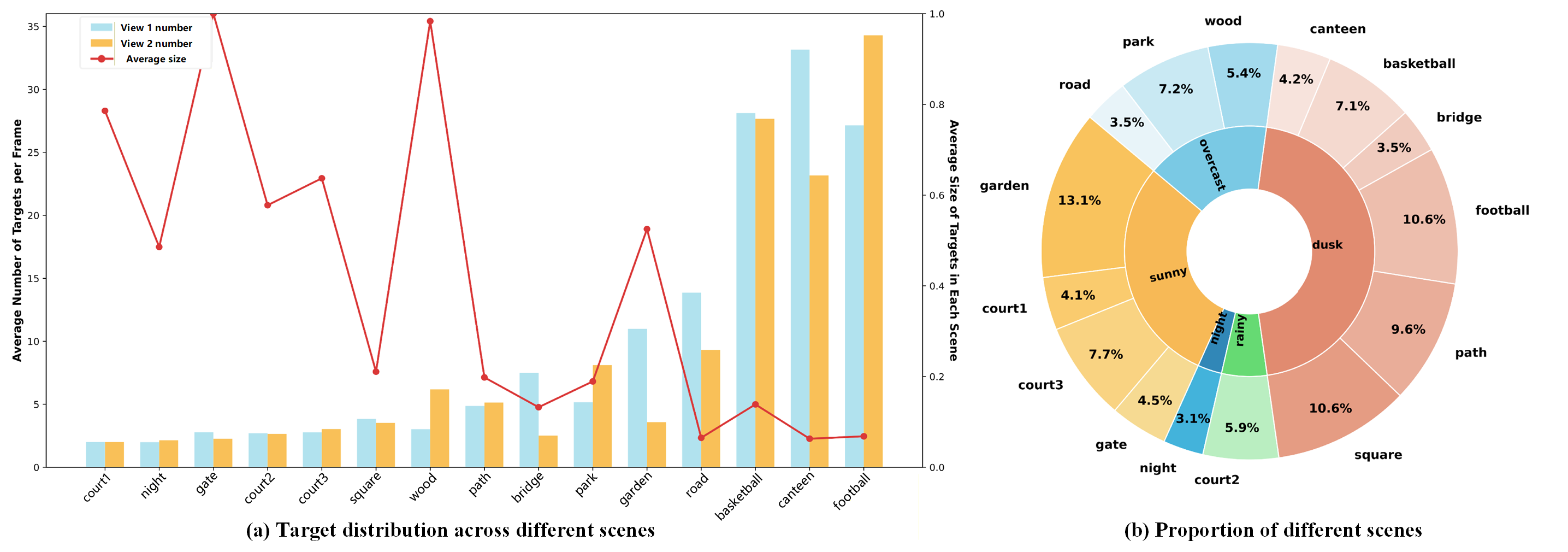}
\vspace{-16pt}

\caption{\textbf{Quantitative analysis of VisionTrack dataset.} (a) The red line represents the average size of each target in different scenes, the bar chart represents the average number of targets per frame. (b) The proportion of different scenes and weather conditions in VisionTrack.}

\label{fig:data_tj}
\end{figure*}

\begin{table*}
\small
  \centering
  \begin{tabular*}{\textwidth}{@{\extracolsep{\fill}} 
                                >{\centering\arraybackslash}p{1.7cm} 
                                >{\centering\arraybackslash}p{1.2cm} 
                                >{\centering\arraybackslash}p{1.2cm}   
                                >{\centering\arraybackslash}p{1.2cm} 
                                >{\centering\arraybackslash}p{1.2cm} 
                                >{\centering\arraybackslash}p{1.2cm} 
                                >{\centering\arraybackslash}p{2.3cm} 
                                >{\centering\arraybackslash}p{1.5cm} 
                                @{}}
    \toprule
    Dataset & Scenes & Sequences & Views & Frames & Boxes & Moving camera  & UAV view  \\
    \midrule
    EPFL & 5 & 4 & 4 & 97K  & 625K & $\times$   & $\times$   \\
    CAMPUS & 4 & 3 & 4 & 83K & 490K & $\times$   & $\times$   \\
    MvMHAT & 1 & 6 & 4 & 31K  & 208K & \checkmark   & $\times$  \\
    WILDTRACK & 1 & 1 & 7 & 3K  & 40K & $\times$   & $\times$  \\
    DIVOTrack & 10 & 20 & 3 & 54K  & 560K & \checkmark   & one  \\
    VisionTrack & 15 & 88 & 2 & 116K & 1176K & \checkmark & two  \\
    \bottomrule
  \end{tabular*}
  \caption{\textbf{Comparison with previous datasets.} VisionTrack significantly surpasses previous datasets in both scene diversity and scale.}
  \vspace{-10pt}
  \label{tab:dataset_tab}
\end{table*}

\section{Related Work}
\subsection{One-stage MCMT Trackers}
Most early MCMT trackers establish relationships among all targets in different views and perform one-stage tracking by constructing a global graph. Chen et al. \cite{equalized} proposed a global graph network that merges SCT and ICT processes. They also introduced a new similarity measurement scheme that balances different similarities in two steps. Liu et al. \cite{liu2017multi} proposed a method using generalized maximum clique optimization to construct a global graph. To better calculate similarity, they used LOMO features and Hankel matrices to represent the appearance and motion features of targets. Makasei et al. \cite{Maksai_2017_ICCV} proposed a global, pattern-based framework for trajectory association depart from the Markovian assumption, enabling both supervised learning from ground truth and unsupervised refinement from raw trajectories. However, due to their inherent limitations and the rapid progress of single-view tracking methods, these approaches were soon surpassed by two-stage methods. 

\subsection{Two-stage MCMT Trackers}
Two-stage tracking approaches decompose MTMC tracking into Single-Camera Tracking (SCT) and Inter-Camera Tracking (ICT). Most current methods perform SCT using off-the-shelf multi-object trackers \cite{wojke2017simple,zhang2022bytetrack,zhang2021fairmot}, while dedicating their contributions to the ICT stage. TRACTA \cite{he2020multi} leverages local trajectories to incorporate richer temporal context and formulates the MTMC data association problem as a local-trajectory to target assignment issue. MIA-NET \cite{liu2023robust} addresses the challenges of multi-camera small object tracking, introduces an inter-camera matching model that employs keypoint mapping. DyGLIP \cite{quach2021dyglip} employs dynamic graphs for link prediction, combining with attention mechanisms to establish accurate data associations among targets. Additionally, some models focus on optimizing the feature representation of targets, making the feature more discriminative across different cameras. MvMHAT \cite{gan2021self}  introduces a self-supervised learning framework, establishing pairwise similarity and triplet transitive similarity for learning data association models in MTMC tracking. Crossmot \cite{hao2024divotrack} introduces both single-view Re-ID embeddings and cross-view Re-ID embeddings, representing target features distinctively in the two tracking steps.
However, these methods typically rely only on intra-view information for SCT and fail to directly exploit multi-view information, which ultimately hinders their performance. 
This observation motivates us to rethink multi-camera tracking from a global perspective, inspiring the design of a global framework that integrates information from all views in a joint tracking process.

\begin{figure*}  
\centering
\includegraphics[width=0.95\linewidth]{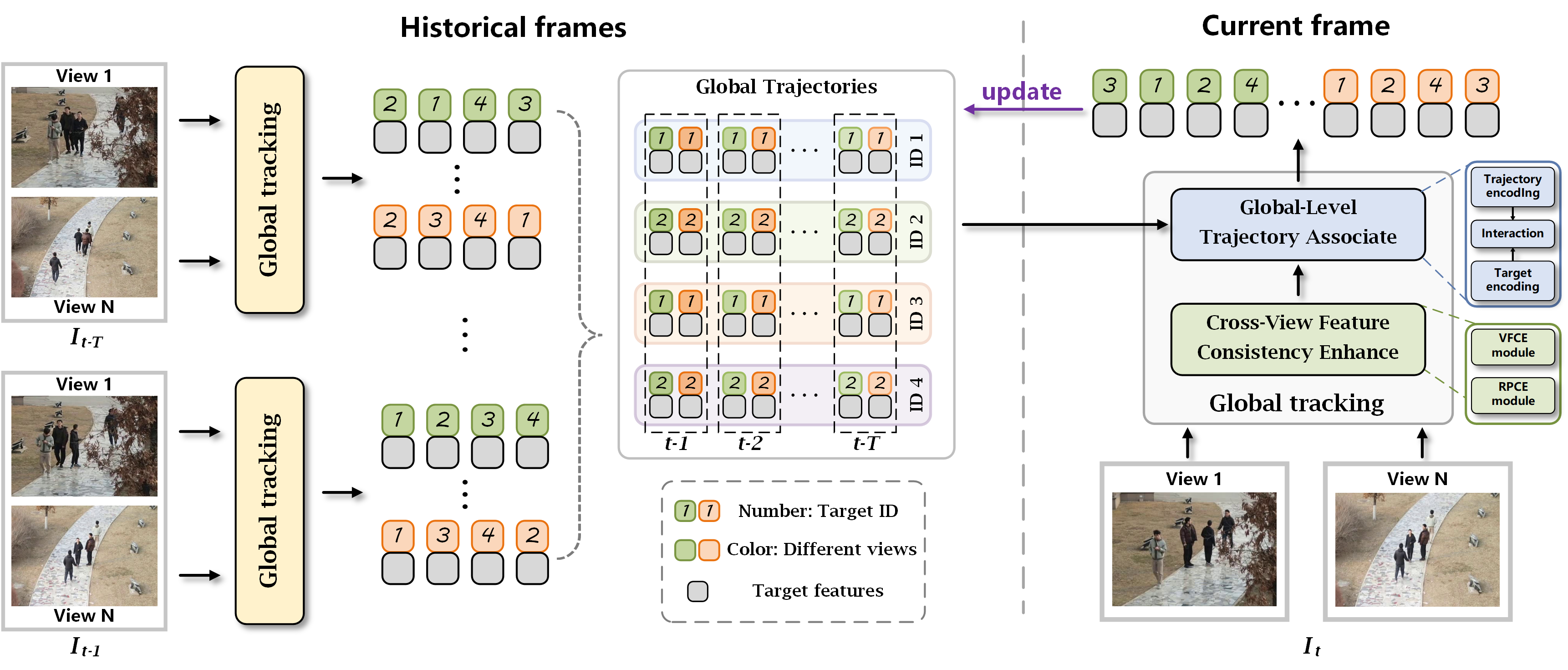}

\caption{\textbf{Overview of the Global MCMT tracking framework(GMT).} GMT is composed of the CFCE module, which enhances intra-trajectory consistency, and the GTA module, which enables trajectory encoding, target encoding and target-trajectory interaction. By encoding the same targets from different views into global trajectories and performing global-level trajectory-target interaction, GMT could directly utilize multi-view information for tracking while avoiding the unnecessary cross-view matching.}
\label{fig:network}
\end{figure*}

\section{VisionTrack Dataset}
\subsection{Contribution of VisionTrack}
Existing MTMC tracking datasets with overlapping fields include: EPFL \cite{fleuret2007multicamera}, CAMPUS \cite{xu2016multi},  WILDTRACK \cite{gan2021self}, MvMHAT \cite{gan2021mvmhat} and DIVOTrack \cite{hao2024divotrack}. It can be observed in \cref{tab:dataset_tab} that these datasets are confined to limited scenarios, exhibit insufficient diversity, and are considerably small in scale. 

To address the limitations and enrich the diversity of MCMT datasets, we construct a novel high-quality, large-scale MCMT dataset, named VisionTrack. As illustrated in \cref{fig:data_tj}, VisionTrack contains 88 sequences across 15 distinct real-world scenarios with diverse environments and varying target densities, offering significantly richer diversity and larger scale.
In addition, unlike other datasets that are captured with fixed ground cameras, we employ two moving UAVs for data collection. We provide more details and visualization about VisionTrack in \textbf{Section \textcolor{red}{B.3} of Appendix.}

\subsection{Data Collection and Annotation}
We captured video data using two drones, both drones were moving during the recording process. All videos were captured at a frame rate of 30 FPS and a resolution of 1920×1080. To ensure dataset diversity, we selected different scenes with varying weather and lighting conditions (shown in \cref{fig:data_tj} (b)). We also captured scenes with varying levels of target density ranging
from sparse large targets to dense small targets (shown in \cref{fig:data_tj} (a)). The dataset is split approximately 1:1 for training and testing.

 Our dataset annotation process was twofold. In the first step, annotators identify and assign unique IDs to the same targets present in the overlapping fields of view of different cameras at time t.
 In the second step, we use the semi-automatic object tracking annotation software, Darklable, to associate the targets at time $t$ with those at time $t-1$ (already annotated) and unify the IDs of the same targets across views based on the results of the first step.

\section{Method}
\label{sec:formatting}

\subsection{Overview}

The overall architecture of GMT is illustrated in \cref{fig:network}. At time $t$, a detector is used to detect all targets from $c$ cameras, yielding a set of all bounding boxes \(B_t=\{B_i\}_{i=1}^N\) and the corresponding target features \(F_t=\{F_i\}_{i=1}^N\). To reduce the feature discrepancies of the same target across different views, we propose a CFCE strategy, composed of VFCE and RPCE modules.
The VFCE module projects $F_t$ from view-centric space to \(F_t^\tau=\{F_i^\tau\}_{i=1}^N\) in trajectory-centric space. 
The RPCE module encodes $B_t$ to relative position features \(F_t^p=\{F_i^p\}_{i=1}^N\). $F_t^\tau$ and $F_t^p$ are concatenated to form the association features \(F_t^{asso}=\{F_i^{asso}\}_{i=1}^N\). 
We perform global-level trajectory-target matching via the GTA module. The global trajectory \(\Gamma=\{\tau_k\}_{k=1}^K\) contains historical information from all views, enabling the direct use of multi-view information during the tracking process. 
The GTA module allows $F_t^{asso}$ to interact with \(\Gamma\) to learn discriminative information of each trajectory, producing the enhanced features $\overline{F}_t^{asso}$. Finally, the GTA module computes the similarity matrix between $\overline{F}_t^{asso}$ and \(\Gamma\), and obtains the tracking result using the Hungarian algorithm.

\subsection{Cross-View Feature Consistency Enhancement}
The purpose of the CFCE module is to enhance global trajectory modeling by reducing the discrepancies among targets within the same trajectory caused by viewpoint differences. Unlike existing methods that rely only on visual information, CFCE introduces an innovative approach that captures the relative spatial relationships between targets,
further improving the accuracy of cross-view matching.

\noindent\textbf{Visual Feature Consistency Enhancement.} The VFCE module aims to transform the target features \(F_t\) from view-centric to trajectory-centric space. We adopt a two-layer MLP as the projection head to project \(F_t\) into \(F_t^\tau\). Following the common approach in cross-view alignment \cite{yu1,shiben1}, we supervise the ID consistency of \(F_t^\tau\) with cross-entropy loss, triplet loss, and center loss. The overall loss of VFCE is formulated as:
\begin{equation}
\mathcal{L}_{\text{VFCE}} = \mathcal{L}_{\text{CE}} +  \mathcal{L}_{\text{Triplet}} +  \mathcal{L}_{\text{Center}},
\end{equation}
where $\mathcal{L}_{\text{CE}}$, $\mathcal{L}_{\text{Triplet}}$, and $\mathcal{L}_{\text{Center}}$ represent the cross-entropy loss, the triplet loss, and the center loss respectively. Each predicted box is assigned the ID of the ground truth box that yields the maximum IoU.


\noindent\textbf{Relative Position Consistency Enhancement. }The RPCE module is motivated by the projection consistency across views \cite{liu2023robust,Chen_2025_CVPR}. Specifically, for images of the same scene captured from different views, if a graph is constructed with all target positions for each view, the resulting graphs from different views can be aligned through affine transformation. Exploiting this consistency of spatial distribution could enable more accurate cross-view association. 

For each individual target in the overlapping field of view, the subgraph formed by its surrounding targets also follows the projection consistency. Therefore, for the $i$-th target, we construct its spatial relation with neighboring targets as:
\begin{equation}
G_{i} = [x_i, y_i , \Delta x_1,\Delta y_1,...,\Delta x_M,\Delta y_M],
\end{equation}
where $x_i, y_i$ denote the normalized center coordinates of the $i$-th target, $\Delta x_j,\Delta y_j$ denote the differences between the center coordinates of the $j$-th neighboring target and $x_i, y_i$. If $G_{i}$ includes neighboring targets that do not appear in the overlapping field, it could introduce noise into cross-view matching. A natural observation is that the closer the cameras are to the scene, the smaller the potential overlapping region becomes in physical space. Motivated by this, we propose a distance-based filtering strategy:
\begin{equation}
r=\left(2 + 2\cdot \mathrm{clip}\!\left(\frac{r_{max} - \sqrt{s_i}}{r_{max} - r_{\min}},\,0,\,1\right)\right)\sqrt{s_i},
\label{eq:einstein}
\end{equation}
where $s_i$ denotes the normalized box size of target, $r_{max}$ and $r_{min}$ are the predefined scaling boundaries, and $\mathrm{clip}(x,a,b)$ denotes the truncation of $x$ to the interval $[a,b]$. When computing $G_{i}$, we only consider neighboring targets within the distance threshold $r$. The scaling in Equation \eqref{eq:einstein} decreases for larger $s_i$, which limits the real-world spatial extent of $r$ and reduces the chance of considering targets outside the overlapping field. We utilize a two-layer MLP to encode \(G_t=\{G_i\}_{i=1}^n\) to $F_t^p$, and supervise this process with triplet loss and center loss:
\begin{equation}
\mathcal{L}_{\text{RPCE}} = \mathcal{L}_{\text{Triplet}} +  \mathcal{L}_{\text{Center}}.
\end{equation}

$F_t^{\tau}\in \mathbb{R}^{N \times d_1}$ and $F_t^{p}\in \mathbb{R}^{N \times d_2}$ are concatenated to construct the association features of all current targets $F_t^{asso}\in \mathbb{R}^{N \times d}$ , where $d=d_1+d_2$. 

\subsection{Global-Level Trajectory Association}
Unlike the two-stage framework that performs intra-view tracking and cross-view association separately, our proposed global framework unifies these two stages into a single global-level trajectory–target association process.
Specifically, we merge the local trajectories of the same target from different views into a unified global trajectory after eliminating their cross-view discrepancies. This allows us to assign consistent target IDs across views before tracking begins, eliminating the need for extra cross-view association. Moreover, leveraging global trajectories enables the tracker to directly exploit the rich visual representations provided by multiple cameras, fully realizing the advantage of multi-view tracking.

\begin{figure}[t]
\centering
\includegraphics[width=0.95\linewidth]{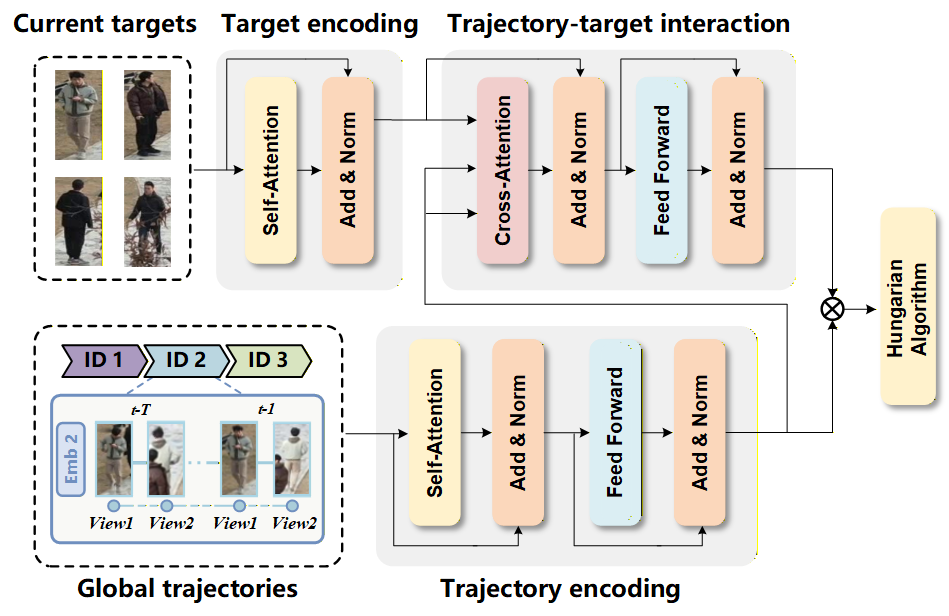}
\caption{\textbf{The structure of the GTA module.} The GTA module implements the global-level trajectory-target association after encoding the global trajectory and the current target features.}
\label{fig:mamba}
\end{figure}

Since the SORT-based \cite{sort,deepsort} trackers used by existing methods rely heavily on spatial consistency and cannot handle multi-view inputs, we construct the GTA module following the DETR-based trackers \cite{zeng2021motr,zhou2022global}. The detailed structure is illustrated in \cref{fig:mamba}.
To enhance trajectory modeling while enabling interactions across trajectories for learning discriminative inter-trajectory cues, we concatenate the features of all detected targets from the most recent $T$ frames as the global trajectory representation, and assign a shared ID embedding to targets within the same trajectory. The resulting feature $\Gamma\in \mathbb{R}^{L \times d}$ is encoded with an encoder layer.  This design is proven more effective than more complex alternatives in \textbf{Section \textcolor{red}{C.2} of the Appendix}. The encoded trajectory features $\overline{\Gamma}$ are then fed into a decoder to interact with $F_t^{asso}$, where each candidate target in $F_t^{asso}$ first interacts with each other to encode their contextual features and subsequently serve as queries to learn discriminative cues between trajectories, producing $\overline{F}_t^{asso}$. 

During inference, GTA computes the similarity matrix $M_{s}\in \mathbb{R}^{N \times L}$ between each target in $\overline{F}_t^{asso}$ and all historical targets in $\overline{\Gamma}$. The similarity to a trajectory is defined as the average similarity to all its associated targets, forming a target–trajectory similarity matrix $M\in \mathbb{R}^{N \times K}$. The Hungarian algorithm is applied to obtain the association results. We utilize a memory module to recover long-term occluded trajectories. Specifically, historical trajectories that have been missing for the past $T$ frames but were observed earlier within a time threshold are stored in the memory bank.
Targets in $\overline{F}_t^{asso}$ that fail to match with $\overline{\Gamma}$ will be matched again with those stored in the memory bank.

During the training stage, both the encoder and the decoder take all target features from $T$ frames as input and perform association for each frame simultaneously. Our training objective is to maximize the log-likelihood between the target-level  similarity matrix $M_{s}$ and the ground-truth matrix. We append an all-zero column to $M_{s}$ as the label for cases where a target in $\overline{\Gamma}$ does not match any target in $\overline{F}_t^{asso}$ and then apply a softmax operation to transform it into a probability distribution $H\in \mathbb{R}^{(N+1) \times L}$:
\begin{equation}
H_{ij} = \frac{e^{M_{ij}}}{e^{M_{i0}} + \sum_{q=1}^{N} 1_{(t_q = t_j \text{ and } c_q = c_j)} e^{M_{iq}}} \label{pythagorean}, 
\end{equation}
where \((t_q = t_j \text{ and } c_q = c_j)\) denotes that the target is detected in time $t_j$ and view $c_j$. 

The final association loss is calculated as follows:
\begin{equation}
X_{ij} = 
\begin{cases} 
1 & \text{if } p_i \leftrightarrow p_j \\
0 & \text{otherwise} ,
\end{cases} \label{pythagorean} 
\end{equation}


\begin{equation}
\begin{split}
\mathcal{L}_{asso} = -\frac {1}{(N+1)} \Bigg(&\sum_{i=1}^{N+1}\Bigg(\sum_{j=1}^{L} X_{ij} \log(H_{ij}) \Bigg)\Bigg).
\end{split}
\end{equation}
where $X\in \mathbb{R}^{(N+1) \times L}$ is the ground-truth matrix, $p_i \leftrightarrow p_j$ denotes that the $j$-th target in $\overline{\Gamma}$ matches the $i$-th label in $H$.

\renewcommand{\arraystretch}{1.15} 
\begin{table*}
\footnotesize
  \centering
  \begin{tabular}{l|cccccc|cccccc}
    \specialrule{0.7pt}{0pt}{0pt}
      \multicolumn{1}{l|}{\multirow{2}{*}{\textbf{Method}}} &
       \multicolumn{6}{c|}{\textbf{VisionTrack}} 
      & \multicolumn{6}{c}{\textbf{DIVOTrack}}
      \rule{0pt}{2.5ex}\\
     \hhline{~------------}
          &CVMA & 
           CVIDF1&
          MOTA & 
          HOTA & 
          IDF1 &
          ASSA& 
          CVMA & 
          CVIDF1 &
          MOTA & 
          HOTA &  
          IDF1 &
          ASSA
          \rule{0pt}{2.5ex}\\
    \hline
    AGW \cite{ye2021AGW} & 70.1 & 68.4 & 77.0 & 59.5 &77.0&57.0& 68.0 &63.9 & 75.9 & 57.6 &68.5&53.1\\
    \rowcolor{tea}CT \cite{ct}& 65.0 & 62.7 & 76.7 & 55.8 & 65.5 & 50.9  &67.6 & 66.5 & 75.6 & 58.2 & 62.6 & 54.4 \\
    MvMHAT$_{prev} $ \cite{gan2021mvmhat} & 39.2 & 53.6 & 76.9 & 56.3 & 66.0 & 51.3 & 64.3 & 57.7 & 78.4 & 60.2 & 70.5 & 55.0  \\
     \rowcolor{tea}CrossMOT \cite{hao2024divotrack}& 64.5 & 64.4 & 75.6 & 54.5 & 65.5 & 49.3 &69.0 & 69.1 & 77.8 & 60.5 & 73.9 & 57.1  \\
     MvMHAT$_{Res} $\cite{MvMHAT++} & 51.1 & 61.3 & 77.0 & 56.0 & 66.4 & 50.8 & 67.5 & 65.3 & 78.9 & 62.0 & 70.1 &51.8\\
     \rowcolor{tea}MvMHAT$_{Trans}$\cite{MvMHAT++}& 68.8 & 70.0 & 77.6 & 59.6 & 72.3 & 57.2 & 69.4 & 68.6 & 78.9 & 64.8 & 74.1 & 56.3\\
          \hline

     GMT & \underline{\textbf{75.2}} & \underline{\textbf{81.3}} & \underline{\textbf{78.0}} & \underline{\textbf{66.2}} & \underline{\textbf{82.1}} &\underline{\textbf{69.4}} & \underline{\textbf{74.5}} & \underline{\textbf{73.2}} & \underline{\textbf{80.5}} & \underline{\textbf{64.5}} & \underline{\textbf{76.7}} &\underline{\textbf{62.1}}\\
     \specialrule{0.6pt}{0pt}{0pt}

  \end{tabular}

      \vspace{+6pt}
  \begin{tabular}{l|cccccc|cccccc}
    \specialrule{0.7pt}{0pt}{0pt}
      \multicolumn{1}{l|}{\multirow{2}{*}{\textbf{Method}}} &
       \multicolumn{6}{c|}{\textbf{WildTrack}}
      & \multicolumn{6}{c}{\textbf{MvMHAT}}
      \rule{0pt}{2.5ex}\\
     \hhline{~------------}
          &CVMA & 
          CVIDF1 &
          MOTA & 
          HOTA & 
          IDF1 &
          ASSA & 
          CVMA & 
        CVIDF1 &
          MOTA & 
          HOTA &  
          IDF1 &
          ASSA 
          \rule{0pt}{2.5ex}\\
    \hline
    AGW \cite{ye2021AGW}& 15.9 & 23.2 & 52.6 & 51.4 & 63.7 & 57.9 &89.6& 90.4 & 89.7 & 71.0 & 73.1 & 60.9\\ 
    \rowcolor{tea}CT \cite{ct} & 17.3 & 27.7 & 53.6 & 51.6 & 46.3 & 57.5 & 48.9 & 56.6 & 93.7 & 74.9 & 84.2 & 66.7\\
    MvMHAT$_{prev} $ \cite{gan2021mvmhat}& 9.4 & 15.9 & 47.4 & 43.4 & 51.6 & 44.1 & 82.4 & 71.8 & 93.1 & 75.7 & 82.8 & 67.5   \\
     \rowcolor{tea}CrossMOT \cite{hao2024divotrack}& 40.4 & 54.8& 55.6 & 49.8 & 63.4 & 52.7  & 90.9 & 86.1 &  95.9 & 78.0 & 90.5 &74.0 \\
     MvMHAT$_{Res} $\cite{MvMHAT++} & - & - & - & - & - & - &89.0 & 85.9 & 92.0 & 82.0 & 86.9 & 70.6 \\
     \rowcolor{tea}MvMHAT$_{Trans}$\cite{MvMHAT++} & - & - & - & - & - & - &89.6 & 88.9 & 91.8 & 82.3 & 87.7 & 71.6\\
     \hline
     GMT & \underline{\textbf{61.7}} & \underline{\textbf{72.0}} & \underline{\textbf{64.9}} & \underline{\textbf{56.1}} & \underline{\textbf{74.1}} & \underline{\textbf{59.1}}  & \underline{\textbf{94.1}} & \underline{\textbf{95.6}} &  \underline{\textbf{96.2}} & \underline{\textbf{87.5}} & \underline{\textbf{95.8}} &\underline{\textbf{85.3}} \\
     \specialrule{0.6pt}{0pt}{0pt}
     
  \end{tabular}

      \vspace{+6pt}
  \begin{tabular}{l|cccccc|cccccc}
    \specialrule{0.7pt}{0pt}{0pt}
      \multicolumn{1}{l|}{\multirow{2}{*}{\textbf{Method}}} &
       \multicolumn{6}{c|}{\textbf{CAMPUS}} 
      & \multicolumn{6}{c}{\textbf{EPFL}}
      \rule{0pt}{2.5ex}\\
     \hhline{~------------}
          &CVMA & 
          CVIDF1 &
          MOTA & 
          HOTA & 
          IDF1 &
          ASSA& 
          CVMA & 
          CVIDF1 &
          MOTA & 
          HOTA &  
          IDF1 &
          ASSA 
          \rule{0pt}{2.5ex}\\
    \hline
    AGW \cite{ye2021AGW} & 51.2 & 50.5 & 58.5 & 41.9 & 50.2 & 33.4 &51.6& 39.9 &70.5 &30.6 & 32.5 & 16.1\\
    \rowcolor{tea}CT \cite{ct} & 58.5 & 47.8 & 59.0 & 42.7 & 52.3 & 34.6 &40.5 & 28.6 & 69.3 & 30.5 & 58.1 & 16.2 \\
    MvMHAT$_{prev} $ \cite{gan2021mvmhat} & 45.7 & 43.6 & 60.0 & 40.6 & 49.7 & 31.4& 29.1 & 20.3 & 77.2 & 22.5 & 21.3 & 9.1\\
     \rowcolor{tea}CrossMOT \cite{hao2024divotrack} & 65.4& 52.4 & \underline{\textbf{72.2}} & 45.9 & 56.6 & 35.8 & 70.8 & 59.1 & 77.6 & 36.2 & 38.2 & 21.3 \\
     MvMHAT$_{Res} $\cite{MvMHAT++}& 50.8 & 45.7 & 66.3 & 36.8 & 48.9 & 34.7 &31.1 & 23.5 & 78.5 & 30.1 & 31.6 & 14.8 \\
\rowcolor{tea}MvMHAT$_{Trans}$\cite{MvMHAT++}& 58.2 & 51.2 & 68.5 & 44.7 & 55.8 & 37.1 &65.9 & 35.0 & 77.6 & 34.6 & 37.7 & 19.8 \\
          \hline
     GMT & \underline{\textbf{66.4}} & \underline{\textbf{66.8}} & 69.5 & \underline{\textbf{52.3}} & \underline{\textbf{69.1}} & \underline{\textbf{48.2}} & \underline{\textbf{73.9}}& \underline{\textbf{66.7}} & \underline{\textbf{78.2}} & \underline{\textbf{51.2}} & \underline{\textbf{67.5}} & \underline{\textbf{42.2}}   \\
     \specialrule{0.6pt}{0pt}{0pt}
     
  \end{tabular}
  \caption{Comparison with state-of-the-art MCMT trackers on VisionTrack, DIVOTrack, WildTrack, MvMHAT, CAMPUS and EPFL. The best result are shown in \underline{\textbf{bold}}.}
  \vspace{-10pt}
  \label{tab:sota}
\end{table*}
\renewcommand{\arraystretch}{1.0} 

\subsection{Training Procedure}

Given that both the RPCE module and the GTA module rely on accurate target localization, we decompose training into two stages. In the first stage, we train the detector and the VFCE module:
\begin{equation}
\mathcal{L}_{\text{stage1}} = \mathcal{L}_{\text{det}} +  \mathcal{L}_{\text{VFCE}},
\end{equation}
we employ DLA-34 \cite{dla} as the backbone and CenterNet \cite{duan2019centernet} as the detector, both are commonly used by previous MCMT trackers. $\mathcal{L}_{\text{det}}$ represents the original loss function of CenterNet.

In the second stage, we train the complete model while excluding the metric learning losses in the VFCE module:
\begin{equation}
\mathcal{L}_{\text{stage2}} = \mathcal{L}_{\text{det}} +  \lambda_1\mathcal{L}_{\text{asso}}+  \lambda_2\mathcal{L}_{\text{VFCE}},
\end{equation}
where $\lambda_1=3$ and $\lambda_2=0.5$. It should be noted that our model could also be trained in one stage, with only slight performance degradation. We provide specific details in \textbf{Section \textcolor{red}{D.6} of the Appendix.}

\section{Experiments}

\subsection{Implementation Details}

 The GMT is trained for 10,000 iterations in each of the two stages with a batch size of 8 on two NVIDIA RTX L40 GPUs. We use the Adam optimizer with a learning rate of $5 \times 10^{-4}$. During inference, the model is evaluated on a single NVIDIA A6000 GPU. The confidence threshold of the target detector is set to 0.525, which is consistent with most previous works.
\subsection{Evaluation Metrics}
We adopt MOTA \cite{MOTA} and HOTA \cite{HOTA}, which jointly evaluate detection and association accuracy, as well as IDF1 \cite{IDF1} and ASSA \cite{HOTA}, which place greater emphasis on identity consistency, as single-view evaluation metrics. To evaluate cross-view tracking performance, we employ the CVIDF1 and the CVMA \cite{CVMHT}.

CVIDF1 is a cross-view version of IDF1, which focuses on measuring ID consistency across all views:
\begin{equation}
CVIDF1 = \frac{2 \times CVIDP \times CVIDR}{CVIDP + CVIDR},
\end{equation}
where CVIDP and CVIDR represent the Cross-View ID Precision and Cross-View ID Recall, respectively.

CVMA is a cross-view version of MOTA and is defined as:
\begin{equation}
CVMA = 1 - \frac{\sum\limits_t \left(m_t + fp_t + 2mm e_t\right)}{\sum\limits_t g_t},
\end{equation}
where ${{m_t}}$ and ${{g_t}}$ represent the number of missed detections and ground truth targets from all views at time $t$,
${f{p_t}}$ and ${mm{e_t}}$ are the counts of false positives and wrong matching pairs, respectively.

\subsection{Comparisons With State-of-The-Art Methods}To more comprehensively evaluate the performance of GMT, we conduct extensive experiments with state-of-the-art MCMT tracking methods on EPFL \cite{fleuret2007multicamera}, CAMPUS \cite{xu2016multi},  WILDTRACK \cite{gan2021self}, MvMHAT \cite{gan2021mvmhat} and DIVOTrack \cite{hao2024divotrack} as well as the proposed VisionTrack dataset. Since most existing methods are evaluated only on specific datasets, we reproduced their results strictly following their original settings. For a fair comparison, we also replaced their detection boxes with those from the detector of GMT. We provide additional metrics in \textbf{Section \textcolor{red}{D.2} of the Appendix.}

\noindent\textbf{VisionTrack. }The comparative results on VisionTrack are shown in \cref{tab:sota}. Our model achieved a performance of 75.2\% CVMA, 81.3\% CVIDF1, 78.0\% MOTA, 66.2\% HOTA, 82.1\% IDF1 and 69.4\% ASSA. Compared with the second-best method, our method achieves substantial improvements in cross-view matching and ID consistency with increases of 5.1\% IDF1, 12.2\% ASSA, 5.1\% CVMA and 11.3\% CVIDF1, effectively demonstrates the strong advantages of the global tracking framework.

\noindent\textbf{CAMPUS \& EPFL. }Earlier methods \cite{hao2024divotrack,gan2021mvmhat} used only parts of each sequence from the CAMPUS and EPFL datasets. This strategy was adopted for convenience in data processing rather than due to data quality issues. To provide a more comprehensive evaluation, we use the complete datasets. Except for achieving the second best MOTA on the CAMPUS dataset, GMT outperforms all existing methods across all other metrics, demonstrating its effectiveness.

\noindent\textbf{Other datasets. }On the DIVOTrack, MVMHAT, and WildTrack datasets, GMT consistently outperforms all existing methods, achieving improvements of 5.1\% CVMA and 4.1\% CVIDF1 over the second-best method on DIVOTrack, 21.3\% CVMA and 17.2\% CVIDF1 on WildTrack, and 3.2\% CVMA and 5.2\% CVIDF1 on MVMHAT.
It is worth noting that on WildTrack, where the number of views and targets are significantly higher than in other datasets, most existing methods exhibit a noticeable decline in cross-view matching performance ( MvMHAT++ cannot be trained on WildTrack due to its high memory consumption ). In contrast, GMT maintains relatively balanced performance, demonstrating that the global tracking framework aligns more naturally with the inherent nature of MCMT tracking.

\begin{figure*}  
\centering
\includegraphics[width=1\linewidth]{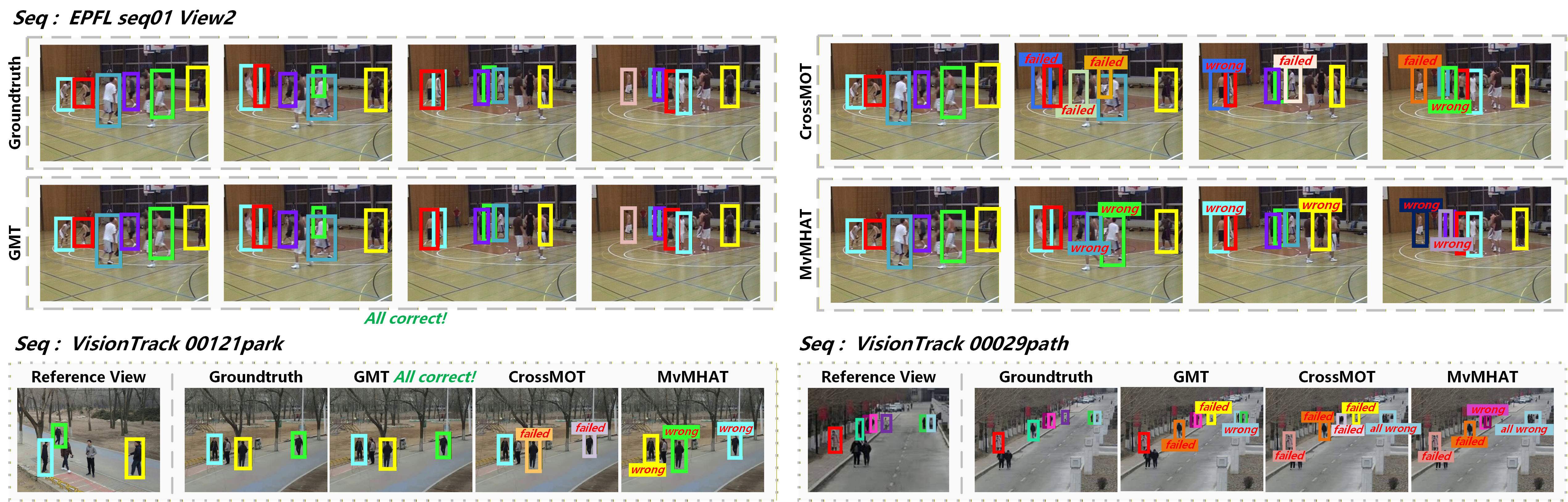}
\vspace{-16pt}
\caption{\textbf{Qualitative comparison} of single-view tracking (\textbf{EPFL 01}) and cross-view matching (\textbf{park, path}) under challenging scenarios. Colors of the target bounding boxes denote distinct target IDs. ‘Fail’ indicates that the tracker fails to maintain tracking of the target and treats it as a newborn target, while ‘Wrong’ denotes an incorrect association with a different target.}
\vspace{-6pt}
\label{fig:keshihua}
\end{figure*}

\subsection{Ablation Study}

\textbf{Cross-View Feature Consistency Enhancement. }The Ablation results of the CFCE module are shown in \cref{tab:ablation}. Since VFCE is also used to adjust the feature dimensions, it cannot be removed directly, \textbf{w/o VFCE} therefore denotes removing the $\mathcal{L}_{\text{VFCE}}$ instead. Introducing VFCE leads to consistent improvements in both intra-view and cross-view tracking performance, yielding gains of 2.8\% CVMA, 3.2\% CVIDF1, and 4.9\% IDF1.
The use of RPCE (\textbf{w/o RPCE}) mainly enhances cross-view association performance, improving CVMA by 1.2\% and CVIDF1 by 1.5\%.
These results indicate that visual features alone are sufficient for single-view tracking, while RPCE fulfills its intended goal of providing complementary cues for cross-view association.
Moreover, removing the $\mathcal{L}_{\text{RPCE}}$ (\textbf{w/o $\mathcal{L}_{\text{RPCE}}$}) and the distance threshold (\textbf{w/o thres}) causes performance degradation, some metrics are even lower than completely removing RPCE, demonstrating that these two designs could effectively filter interference and enhance overall performance.

\noindent\textbf{Effect of Global Trajectories. }
To verify the impact of using global trajectories, we independently evaluate GMT on each view of the VisionTrack and report the averaged results across views as the final tracking performance, as shown by the \textbf{Local} results in \cref{tab:singleview}. Utilizing global trajectories brings improvements of 4.0\% IDF1 and 4.5\% ASSA.
Moreover, we evaluated other DETR-based single-view trackers with architectures similar to GMT on VisionTrack. Compared to these methods, GMT achieves improvements of 4.8\% HOTA, 6.2\% IDF1, and 6.3\% ASSA with even worse detection (i.e. lower MOTP).
Another advantage of global trajectories is that they contain richer temporal contextual information. We provide experiments in \textbf{Section \textcolor{red}{D.5} of the Appendix}. Our proposed global tracking framework outperforms the baseline methods even with limited temporal context. 
These results demonstrate the effectiveness of our core idea of directly leveraging multi-view information during tracking.
\begin{table}
  \centering
  \footnotesize
  \begin{tabular}{l|cccccc}
    \specialrule{0.6pt}{0pt}{0pt}
          \textbf{Method}&CVMA&CVIDF1&
          MOTA &  
           HOTA&
          IDF1\\
    \hline
    w/o VFCE  & 72.4 & 78.1 & 76.9 & 62.0 &77.2\\
    w/o RPCE & 74.0 & 79.8 & 77.2 & 64.7 & 68.9 \\ 
     - w/o $\mathcal{L}_{\text{RPCE}}$ & 74.3 & 80.8 & 77.5 & 64.9 & 81.9  \\
     - w/o thres &72.3 &78.2 & 76.4 &61.5 &76.3 \\

     \rowcolor{tea}GMT & \textbf{75.2} &\textbf{81.3} & \textbf{78.0} &\textbf{66.2} & \textbf{82.1} \\
     \specialrule{0.6pt}{0pt}{0pt}
  \end{tabular}
  \vspace{-4pt}
  \caption{Ablation results of the CFCE module.}
  \vspace{-6pt}
  \label{tab:ablation}
\end{table}

\begin{table}
  \centering
  \footnotesize
  \begin{tabular}{l|ccccc}
    \specialrule{0.6pt}{0pt}{0pt}
          \textbf{Method}&
          MOTA & 
          MOTP &
          HOTA & 
          IDF1 &
          ASSA
          \\
    \hline
    MOTR \cite{zeng2021motr} & 74.9 & \textbf{81.1} & 58.4 & 69.3 & 58.3\\
    MOTRv2 \cite{MOTRV2}  & 75.1& 80.8 & 59.8 & 72.1 & 61.7 \\
    GTR  \cite{zhou2022global} & 75.4 & 80.5 & 61.4 & 75.9 & 63.1  \\
     Local & 76.8 & 79.5 & 62.6 & 78.1 & 64.9  \\
     \rowcolor{tea}GMT & \textbf{78.0} & 79.6 & \textbf{66.2} & \textbf{82.1} & \textbf{69.4}   \\
     \specialrule{0.6pt}{0pt}{0pt}
  \end{tabular}
  \vspace{-4pt}
  \caption{Ablation results of the global trajectories.}
  \vspace{-6pt}
  \label{tab:singleview}
\end{table}

\section{Qualitative Comparison}
\cref{fig:keshihua} presents the qualitative comparison between our GMT, CrossMOT \cite{hao2024divotrack}, and the Transformer-based MVMHAT++ \cite{MvMHAT++}. For a more intuitive demonstration, we only labeled the failure cases. The \textbf{EPFL 01} exhibits large view variations, making it suitable for evaluating single-view tracking performance. GMT outperforms baseline methods under conditions with similar targets and severe occlusions. In \textbf{park} and \textbf{path} scenes, all targets are visible in both views, making them suitable for evaluating cross-view matching. GMT achieves the best cross-view matching results under conditions with similar small targets. We provide more visualization in \textbf{Section \textcolor{red}{E} of the Appendix}.


\section{Model Efficiency and Convergence}
\begin{figure}  
\centering
 \includegraphics[width=0.95\linewidth]{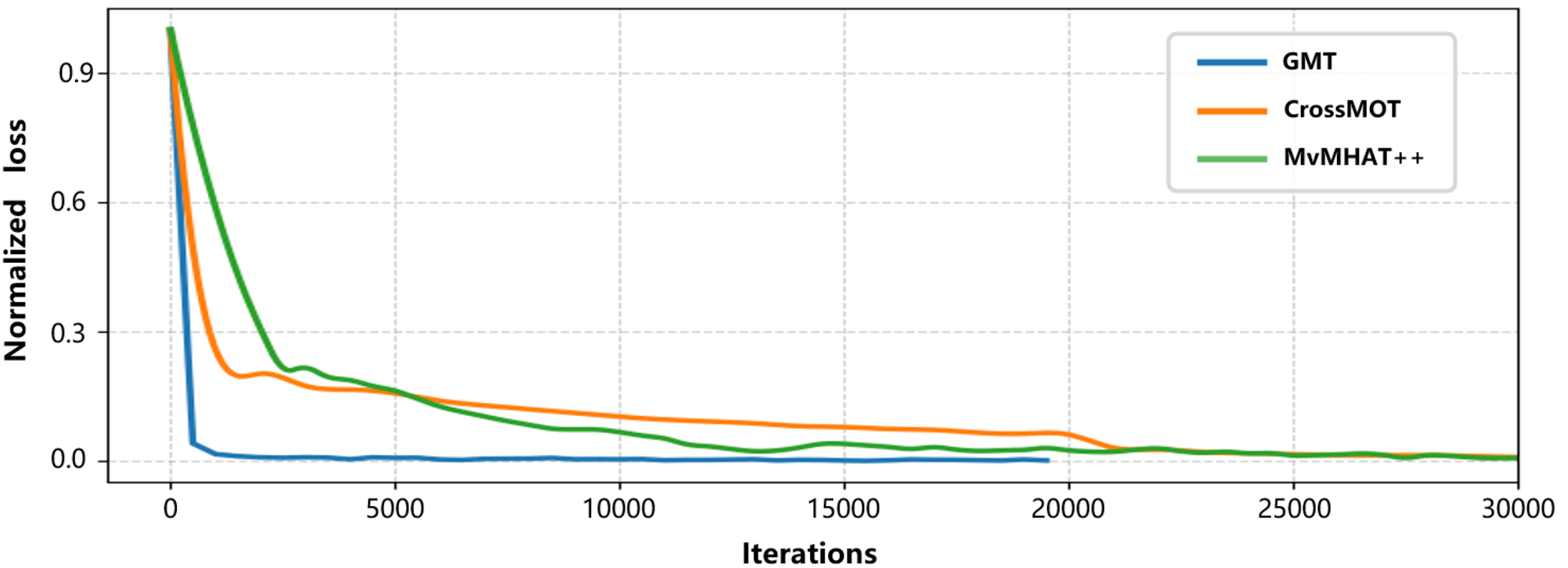}
\vspace{-7pt}
\caption{Comparison of convergence conditions.}
\vspace{-8pt}
\label{fig:loss}
\end{figure}

\cref{tab:param} presents the comparison of model complexity between GMT and the baseline methods. We integrate CenterNet \cite{duan2019centernet} to MvMHAT++ for a fair comparison. GMT achieves superior performance while remaining comparable to existing methods in terms of speed and complexity. Moreover, as shown in \cref{fig:loss}, GMT converges significantly faster benefiting from its concise architecture, requiring only 7.3 hours of training.

\section{Conclusion}
In this work, we propose a novel global tracking framework GMT, which addresses the limitation of insufficient multi-view information utilization in existing two-stage MCMT tracking  framework. Unlike the two-stage framework that performs tracking independently within each view, GMT encodes the same target across different views as a global trajectory and directly leverages multi-view information during tracking by performing global-level trajectory–target matching. Experimental results show that GMT achieves significant performance improvements over two-stage methods. In addition, to address the limitations of existing datasets that are small and lacking diversity, we introduce VisionTrack, a novel large-scale, high-quality MCMT dataset that offers considerably greater diversity.
We hope that our proposed framework and the dataset will bring new vitality together to the field of MCMT tracking.

\begin{table}
  \centering
  \footnotesize
  \begin{tabular}{l|ccccc}
    \specialrule{0.6pt}{0pt}{0pt}
          \textbf{Method}&
          Params & 
          Flops &
          Speed & 
          Training time
          \\
    \hline
    MVMHAT++ \cite{MvMHAT++} & 22.9M & 106.7G & 18.7fps & 11h \\
    CrossMOT \cite{hao2024divotrack}  &24.3M & 227.1G & 14.2fps & 16.5h  \\
    \rowcolor{tea}GMT  & \textbf{30.0M} & \textbf{227.6G} & \textbf{14.6fps} & \textbf{7.3h}   \\
     \specialrule{0.6pt}{0pt}{0pt}
  \end{tabular}
\vspace{-4pt}
  \caption{Comparison of parameter and speed.}
    \vspace{-8pt}
  \label{tab:param}
\end{table}

{
    \small
    \bibliographystyle{ieeenat_fullname}
    \bibliography{main}
}


\end{document}